\documentclass{article}
\pdfoutput=1  

\usepackage{cite} 
\usepackage{graphicx}

\newcommand\Rey{\mbox{\textit{Re}}}  
\newcommand\Wom{\mbox{\textit{W}}}  

\newcommand{\eq}[1]{Eq.~\ref{eq.#1}} 

\newcommand{\fig}[1]{Fig.~\ref{fig.#1}}

\newcommand{\tbl}[1]{Table~\ref{table.#1}}

\newcommand{\eqlabel}[1]{\label{eq.#1}}
\newcommand{\figlabel}[1]{\label{fig.#1}}
\newcommand{\tbllabel}[1]{\label{table.#1}}

\newcommand{\figwidth}{3.5in}

\newcommand{\scenario}[1]{{\sc #1}}

\newcommand{\BoltzmannConstant}{\ensuremath{k_B}}

\newcommand{\density}{\ensuremath{\rho}}
\newcommand{\viscosity}{\ensuremath{\eta}}
\newcommand{\viscosityKinematic}{\ensuremath{\nu}}  
\newcommand{\Tbody}{\ensuremath{T_{\mbox{\scriptsize body}}}}

\newcommand{\Pfluid}{\ensuremath{P_{\mbox{\scriptsize fluid}}}}
\newcommand{\Pinternal}{\ensuremath{P_{\mbox{\scriptsize internal}}}}
\newcommand{\Pfriction}{\ensuremath{P_{\mbox{\scriptsize friction}}}}
\newcommand{\kfriction}{\ensuremath{k_{\mbox{\scriptsize friction}}}}

\newcommand{\geometryFluid}{\ensuremath{h_{\mbox{\scriptsize fluid}}}}
\newcommand{\geometryInternal}{\ensuremath{h_{\mbox{\scriptsize internal}}}}
\newcommand{\geometryLocomotion}{\ensuremath{h_{\mbox{\scriptsize loc}}}}

\newcommand{\dC}{\ensuremath{d_{\mbox{\scriptsize 0}}}}
\newcommand{\vC}{\ensuremath{v_{\mbox{\scriptsize 0}}}}

\newcommand{\vTread}{\ensuremath{v_{\mbox{\scriptsize tread}}}}

\newcommand{\aSphere}{\ensuremath{a_{\mbox{\scriptsize sphere}}}}

\newcommand{\meter}{\mbox{m}}

\newcommand{\millimeter}{\mbox{mm}}
\newcommand{\micron}{\mbox{$\mu$m}}
\newcommand{\nanometer}{\mbox{nm}}

\newcommand{\second}{\mbox{s}}
\newcommand{\millisecond}{\mbox{ms}}

\newcommand{\kg}{\mbox{kg}}
\newcommand{\picowatt}{\mbox{pW}}
\newcommand{\joule}{\mbox{J}}
\newcommand{\Kelvin}{\mbox{K}}
\newcommand{\Pascal}{\mbox{Pa}}

\newcommand{\ms}{\meter^2/\second}

\title{Energy Dissipation by Metamorphic Micro-Robots in Viscous Fluids}
\author{Tad Hogg\\Institute for Molecular Manufacturing\\Palo Alto, CA}

\begin{document}
\maketitle

\begin{abstract}

Microscopic robots could perform tasks with high spatial precision, such as acting on precisely-targeted cells in biological tissues. Some tasks may benefit from robots that change shape, such as elongating to improve chemical gradient sensing or contracting to squeeze through narrow channels. This paper evaluates the energy dissipation for shape-changing (i.e., metamorphic) robots whose size is comparable to bacteria. Unlike larger robots, surface forces dominate the dissipation. Theoretical  estimates indicate that the power likely to be available to the robots, as determined by previous studies, is sufficient to change shape fairly rapidly even in highly-viscous biological fluids. Achieving this performance will require significant improvements in manufacturing and material properties compared to current micromachines.
Furthermore, optimally varying the speed of shape change only slightly reduces energy use compared to uniform speed, thereby simplifying robot controllers.

\end{abstract}

\section{Introduction}

Metamorphic robots~\cite{arbuckle04,castano00,goldstein09,rus99,salemi01,yim00} can change their shape to operate effectively in variable unstructured environments, e.g., for searching collapsed buildings.
Extending this concept, ongoing progress in reducing robot sizes toward that of biological cells~\cite{sitti15} raises the possibility of microscopic metamorphic robots.
Such robots could be particularly useful in medical applications~\cite{freitas99} by operating in biological fluids.
Examples include robots matching the shape of cells to provide temporary scaffolds during tissue repair, and extending probes into spaces, e.g., between or into cells, that are too small or fragile for the whole robot to enter.

Microscopic organisms and robots face significantly different operating constraints than larger ones~\cite{purcell77}. In particular, their high surface to volume ratio means surface forces, such as viscous drag and internal friction, dominate the behavior of microscopic robots. Surface forces are especially significant for shape change, which involves external surfaces moving through the surrounding fluid and internal surfaces moving past each other to actuate the shape change. Moreover, the need to scavenge power from their environment, rather than relying on tethered power, limits the power available to the robots.
Thus, in addition to the challenges of manufacturing and operating microscopic robots in general, power could constrain how rapidly metamorphic robots change shape.

This paper evaluates power required for microscopic metamorphic robots to change shape in biological fluids, including the effects of both external viscous drag and internal actuator friction. 
This contrasts with one application -- locomotion -- where periodic shape changes have been extensively studied~\cite{lauga09} by focussing almost exclusively on external viscous drag.
Specifically, the remainder of this paper discusses power dissipation for microscopic metamorphic robots, and then evaluates the dissipation for three prototypical examples: a robot expanding its length, a robot extending a thin probe and robots coming together to form an aggregate structure.

\section{Power Dissipation During Shape Change}

In biological tissues, viscosity varies by orders of magnitude while density is roughly the same. Thus as example scenarios we consider a wide range of viscosity but constant density.
These scenarios are changing shape in 1) a low viscosity fluid comparable to water and blood, and 2) a high viscosity fluid $10^4$ times more viscous than water, which is typical for mucus or cell cytoplasm~\cite{freitas99}. 
\tbl{scenarios} gives the fluid parameters.

\begin{table}
\centering
\begin{tabular}{lccc}
scenario				&	&\scenario{low}		& \scenario{high} \\
\hline 
density		& $\density$	&$1000 \,\kg/\meter^3$	&$1000 \,\kg/\meter^3$\\
viscosity		& $\viscosity$	&$10^{-3}\,\Pascal \cdot \second$	&  $10\,\Pascal \cdot \second$\\
kinematic viscosity		& $\viscosityKinematic =\viscosity/\density$ & $10^{-6}\,\ms$ & $10^{-2}\,\ms$\\
\end{tabular}
\caption{Parameters for biological fluids similar to water and cell cytoplasm.
}\tbllabel{scenarios}
\end{table}

Power use depends on the robot's geometry, the speed with which it changes shape, the viscosity of the fluid it operates in and the friction of its internal mechanisms. 
To isolate the effect of changing geometry, we define characteristic robot size ($\dC$) and speed of its shape change ($\vC$). We take these values to be fixed for a particular scenario, thereby separating effects of shape from those of overall size and speed.

The definitions of these characteristic quantities are somewhat arbitrary. For example, the size of a spherical robot could be its radius or diameter. Moreover, a robot may change size and have parts moving at different speeds as it reconfigures. Thus $\vC$ could be the maximum or average speed of all moving robot parts. With a specific choice of these characteristic quantities, the analysis presented here isolates the effects of shape by defining dimensionless geometric factors relating power dissipation to shape. The dependence of the results on $\dC$ and $\vC$ indicate how power varies with respect to overall size and speed.

The following analysis uses the characteristic size and speed
to determine the suitability of simplifying approximations. This is suitable provided all sizes and speeds during the reconfiguration remain close enough to these characteristic values for the approximations to remain valid. Thus, for instance, we do not consider robots that change shape by orders of magnitude in size or have similarly large variations in the maximum speed of moving parts during the change.

\subsection{External Power Dissipation}

To determine power dissipated in the fluid surrounding the robot as it changes shape, we make two approximations to the fluid flow. First, for small objects moving relatively slowly in fluid, viscous forces are significantly larger than inertial forces. This leads to smooth, laminar motion called Stokes flow~\cite{kim05,happel83,white05}.
This approximation is accurate when the flow's Reynolds number
\begin{equation}
\Rey=\frac{\vC \dC}{\viscosityKinematic}
\end{equation}
is small, where $\viscosityKinematic$ is the fluid's kinematic viscosity.

The second approximation is the robot changes shape slowly enough that the fluid motion at each time is close to the static flow associated with the instantaneous geometry~\cite{kim05}. This quasi-static approximation requires a small value of the Womersley number
\begin{equation}
\Wom=\frac{\dC}{\sqrt{\viscosityKinematic T}}
\end{equation}
where $T$ is the time over which the geometry changes.

Since characteristic size and speed are somewhat arbitrary, so are the values of \Rey\ and \Wom. Thus their order of magnitude rather than precise values provide a guide to the validity of the fluid approximations used here.

These two approximations give a simple scaling of energy dissipation with size and speed, allowing a numerical solution with one choice of parameters to also apply to other cases with small Reynolds and Womersley numbers. 
Specifically, the power dissipated has the form
\begin{equation}\eqlabel{Pfluid}
\Pfluid = \geometryFluid \viscosity \dC \vC^2
\end{equation}
where $\viscosity$ is the fluid viscosity
and $\geometryFluid$ is a dimensionless geometric coefficient depending on the robot's shape and how rapidly it changes compared to $\vC$. 
Thus, for given choices of $\dC$ and $\vC$, $\geometryFluid$ determines how power dissipation varies as the robot changes shape.

\subsection{Internal Power Dissipation}

Internal dissipation arises from the robot's actuator mechanisms. Friction in the actuators is the major source of internal dissipation directly related to a robot's shape change.
Friction is a complicated property, depending on surface microstructure, lubrication, applied forces and operating speeds~\cite{krim02,vanossi13}.
Robots made of precisely structured, stiff materials could, in theory, experience much less friction than current micromachines.
Specifically, phonon scattering dominates the friction of smooth surfaces separated by atomic-scale distances moving past each other at speed $v$ well below that of sound in the material~\cite{drexler92}. This leads to power dissipation of the form
\begin{equation}\eqlabel{friction}
\Pfriction = \kfriction S v^2
\end{equation}
where $S$ is the area of the moving surfaces and $\kfriction$ is a constant depending on the materials and their spacing. 
Thus, unlike the complexity of friction in micromachines with poorly defined geometry at nanometer scales, friction in smooth nanoscale surfaces has a simple dependence on area and speed.
In particular, friction depends on the contact area $S$, in contrast to macroscopic systems with friction independent of the area~\cite{cumings00}.

For stiff materials, theoretical estimates of $\kfriction$ give values somewhat less than $10^3\,\kg/(\meter^2 \second)$~\cite{drexler92,freitas99}. We use this upper bound as a conservative estimate of internal friction for hypothetical actuators with stiff, atomically-flat sliding surfaces, i.e., we take
\begin{equation}\eqlabel{k friction}
\kfriction = 1000\,\kg/(\meter^2 \second)
\end{equation}

For comparison, the viscous friction for flat surfaces moving with relative speed $v$ while separated by a layer of fluid with viscosity $\viscosity$ and thickness $d$ is $\viscosity S v/d$. This dissipates power $(\viscosity/d) S v^2$, so $\kfriction=\viscosity/d$. Extrapolating to distances of a few tenths of a nanometer gives $\kfriction \approx 10^7\,\kg/(\meter^2 \second)$ for fluids with viscosity similar to water.
A similar value for $\kfriction$ comes from measured drag of micromachine rotors~\cite{chan11}.
More precisely defined micron-scale surfaces can have lower friction~\cite{yang13}, but still significantly larger than the value of $\kfriction$ of \eq{k friction}.
Thus the internal dissipation we consider, while achievable in principle, is considerably smaller that of current micromachines.

For the small robots considered here, sliding surfaces are a relatively simple structure. As shown below, the internal dissipation from sliding is relatively small compared to dissipation due to viscosity in the surrounding fluid, provided smooth materials with low values of $\kfriction$ become available.
On the other hand, if improved material fabrication can only reduce friction to values an order of magnitude or so larger than that of \eq{k friction}, alternate designs might avoid having internal dissipation dominate the total power use. Examples of such designs include rolling bearings between surfaces or lubricants found in larger-size robots. Another possibility is using flexure joints, which do not involve sliding friction.

We consider the situation where sliding friction dominates internal power dissipation. Other power uses, such as for the control circuits or losses in distributing power to the actuators, are not considered here since these forms of dissipation are not directly related to the changing shape. Thus we take $\Pinternal \approx \Pfriction$. With \eq{friction}, internal dissipation then has the form
\begin{equation}\eqlabel{Pinternal}
\Pinternal = \geometryInternal \kfriction \dC^2 \vC^2
\end{equation}
where $\dC^2$ gives a characteristic area and $\geometryInternal$ is a dimensionless geometric coefficient for internal dissipation.

\subsection{Energy Dissipated to Change Shape}

The energy used to change shape is in time $T$ is
\begin{equation}\eqlabel{energy}
E = \int_0^T P(t) dt
\end{equation}
where
\begin{equation}\eqlabel{power}
P(t) = \Pfluid(t)+\Pinternal(t)
\end{equation}
is the power dissipated at time $t$. The power varies with the robot's shape through the geometric factors $\geometryFluid$ and $ \geometryInternal$ in \eq{Pfluid} and \eq{Pinternal}, respectively.

As an example, suppose the shape change involves a single degree of freedom, specified by a parameter $f(t)$ monotonically increasing from $f=0$ at $t=0$ to $f=F$ at time $T$.
Uniform motion in terms of this parameter corresponds to $f(t)=(F/T)t$.

Both external and internal power are proportional to the speed squared, so the geometric factors $\geometryFluid$ and $ \geometryInternal$  are proportional to $\dot{f}^2$, where $\dot{f}\equiv df/dt$. Thus \eq{power} has the form $P(t) = k(f) \dot{f}^2$ where $k(f)$ is a proportionality factor depending on the robot's configuration (through the parameter $f$), fluid viscosity \viscosity, and \kfriction.

One application of this formulation is determining how to vary the speed of the shape change to minimize the energy dissipated. This amounts to finding the function $f(t)$ that minimizes \eq{energy} subject to the conditions $f(0)=0$ and $f(T)=F$. The Euler-Lagrange equation determines the minimizing choice of $f(t)$ as the one giving constant power use throughout the shape change. Thus the optimal rate to change the parameter is $\dot{f} \propto 1/\sqrt{k(f)}$. That is, the robot should change more slowly in configurations with larger energy dissipation in inverse proportion to the square root of that variation.

In the examples of shape change discussed below, dissipation has only a modest dependence on the changing robot shape. Thus energy-minimizing change has relatively little variation in $\dot{f}$ so that uniform motion is close to optimal. This is a useful practical observation: in these cases, there is little benefit from an elaborate control program to minimize energy instead of moving actuators uniformly. Since microscopic robots have limited computational capability, this observation allows using simple control algorithms without sacrificing much energy efficiency.

\section{Linearly Expanding Robot}

Example applications for microscopic metamorphic robots are 1) using a compact shape to travel through the bloodstream and expanding at an injury location to aid repair~\cite{freitas00}, 2) temporarily increasing length to improve detection of chemical gradients~\cite{dusenbery98a}, 3) changing shape to alter effects of fluid drag and Brownian motion~\cite{dusenbery09}, 
and, 4) forming close contacts with cells or other robots. This section examines an illustrative example of these applications: a robot that changes its length.

\subsection{Geometry}

\begin{figure}
\centering 
\begin{tabular}{c}
 \includegraphics[width=\figwidth]{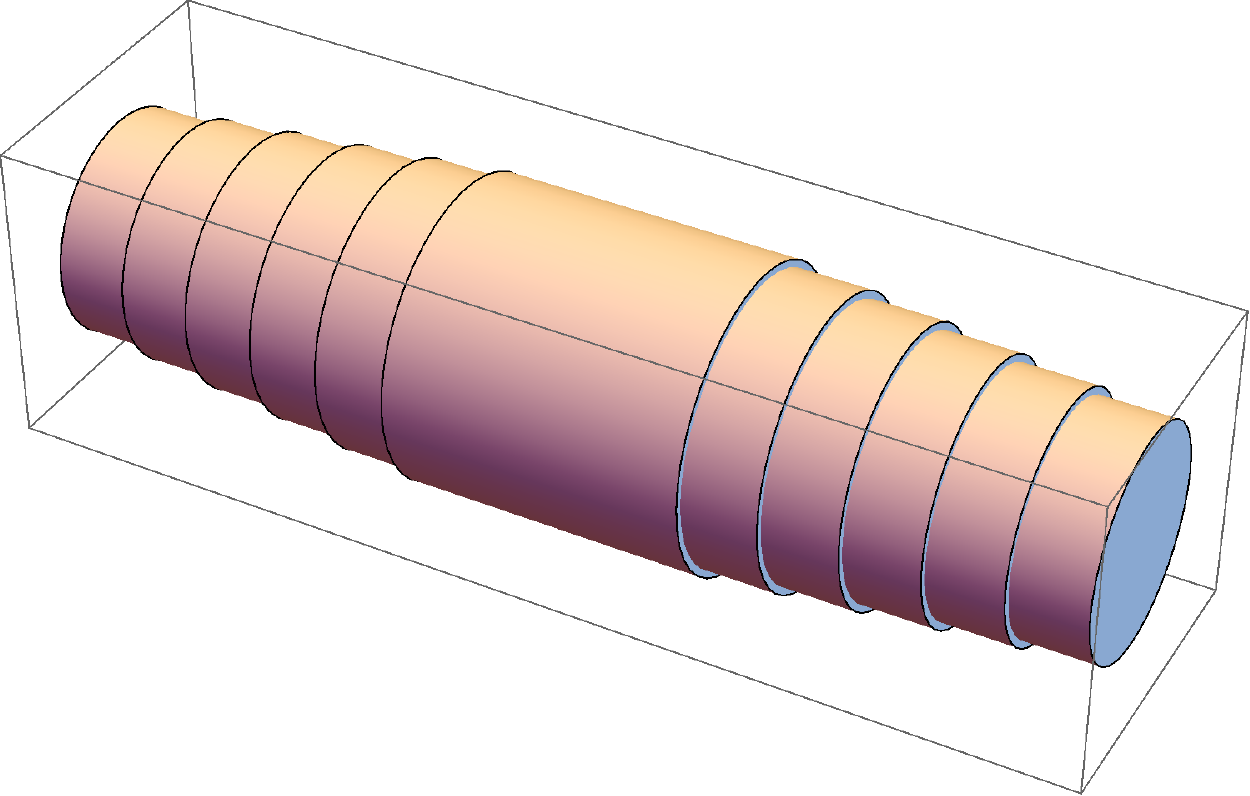}\\
 (a)\\
 \\
  \includegraphics[width=\figwidth]{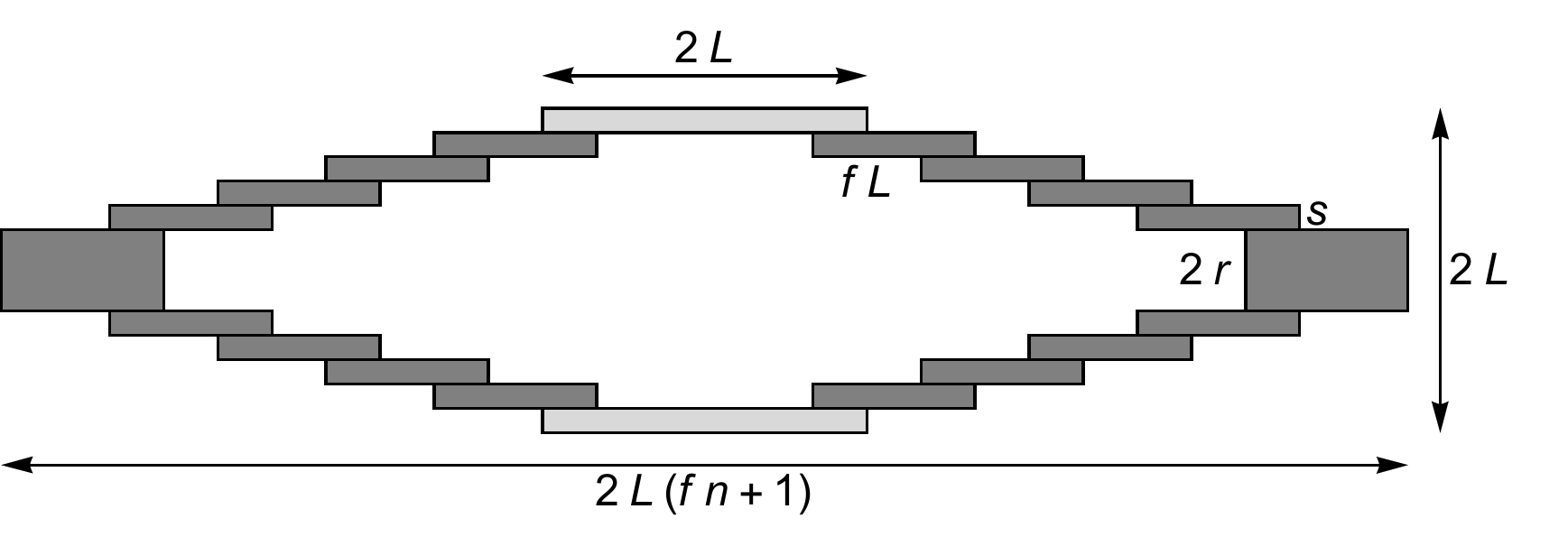}\\
  (b)
\end{tabular}
\caption{Expanding robot with $n=5$ internal segments on each side, with each internal segment extended a fraction $f$ of its full length. (a) Telescoping cylinders with $f=0.5$. (b) Cross section shown with thicker shells than parameters used for evaluating behavior (given in \tbl{expanding}).}\figlabel{extending}
\end{figure}

Consider a metamorphic robot consisting of telescoping segments, allowing it to expand in length, as shown in \fig{extending}. The robot consists of an outer cylinder, of length $2L$ and radius $L$. The robot expands from both ends of the outer cylinder, with each side having $n$ internal segments, each of length $L$. The innermost segment has radius $r=L - n s$ and the rest are shells of thickness $s$.

We suppose all segments extend simultaneously, with all segments continuing to move until the robot reaches its final configuration. Thus a single parameter specifies the robot shape, namely the fraction of extension for each segment, $f$.
With uniform expansion from $f=0$ to $f=F$ in time $T$, $\dot{f}=F/T$ and
convenient choices for characteristic size and speed are the segment length and speed of the ends of the expanding robot:
\begin{eqnarray}
\dC &= & L \\
\vC &=& n L \dot{f}
\end{eqnarray}

\subsection{Behavior}

\begin{figure}
\centering 
 \includegraphics[width=\figwidth]{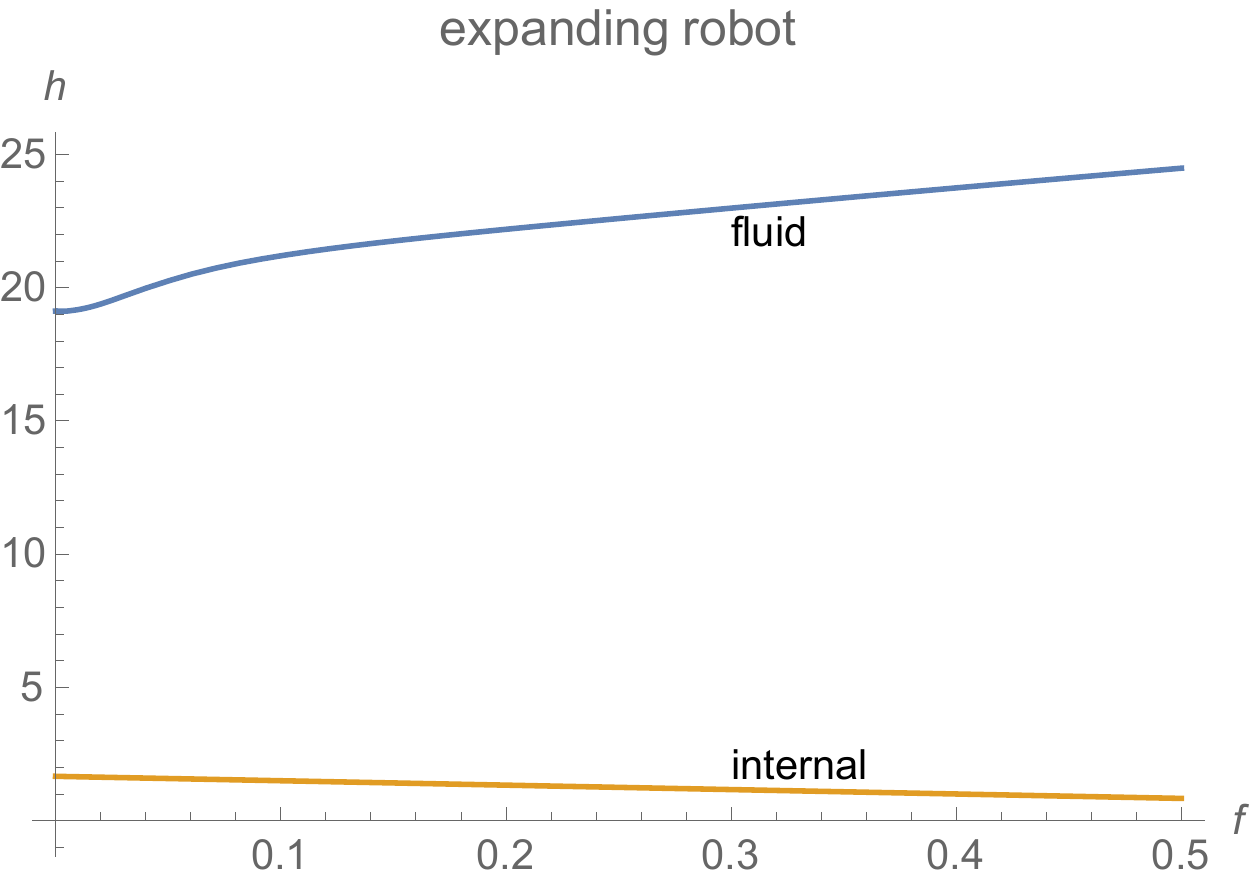}
\caption{Coefficients for viscous drag, $\geometryFluid$, and internal dissipation, $\geometryInternal$, vs.~extent of an expanding robot with $n=5$ segments on each side and ratio of innermost to outer radius $r/L=0.75$.}\figlabel{extending h}
\end{figure}

To evaluate fluid drag we simplify the geometry to truncated cones extending from each side of a cylinder of radius $L$ and length $2L$. The end of the cones move at speed $n L \dot{f}$, and the speed along the side interpolates linearly from 0 at the base to $n L \dot{f}$ at its end. This simplification corresponds to the robot having a thin elastic outer layer covering the actuating shells.
Using this geometry, we solve the steady-state Stokes flow with specific choices of size and speed to determine power dissipation as a function of extension. Dividing result by $\dC \viscosity \vC^2$ gives $\geometryFluid$, shown in \fig{extending h}, which does not depend on the size and speed used to solve the fluid flow.

Evaluating internal power dissipation requires the sliding surface area and speed. The sliding surface area between segments $i$ and $i-1$ is $S_i= 2\pi(r+(n-i)s) L (1-f)$. The surfaces slide past each other with relative speed $v_i-v_{i-1}=L \dot{f}$.
Internal power use arises from the sliding friction between adjacent shells (\eq{friction}):
\begin{equation}
\Pinternal = 2 \sum_{i=2}^n \kfriction S_i (L \dot{f})^2
\end{equation}
with the factor of 2 from the segments expanding in the opposite direction.
With \eq{Pinternal}, this gives
\begin{equation}
\geometryInternal = 2\pi (1-f) \frac{n-1}{n^3} \left((n-2)+(n+2)\frac{r}{L} \right)
\end{equation}
Thus $\geometryInternal$ decreases as the robot expands, i.e., as $f$ increases, since each segment is in contact with less of its neighboring segments. \fig{extending h} shows the behavior of $\geometryInternal$.

With a fixed ratio between radius of innermost and outermost segments and fixed expansion speed of the ends $\vC$, the dissipation from sliding friction decreases with $n$. I.e., using a larger number of thinner segments requires less power than a small number of thicker segments. Thus, the increase in sliding areas due to having more segments is more than offset by the slower relative speeds between neighboring segments.
This discussion does not include losses in the actuators: more segments require more actuators; and there is a minimum feasible thickness to the shells, e.g., to maintain required stiffness and have room for any required transmission of signals or power through the segment walls.

\subsection{Example}

\tbl{expanding} illustrates the behavior for the two scenarios of \tbl{scenarios}. The Reynolds and Womersley numbers are small in both cases, indicating that quasi-static Stokes flow is an adequate approximation for the fluid behavior.

Combined with the geometric coefficients shown in \fig{extending h}, these values indicate viscous drag is the dominant contribution to overall power use, and the power required for extending the robot in the times indicated is well below $1\,\picowatt$. This is well within the power likely available to robots of this size, e.g., from chemical energy~\cite{freitas99,hogg10}.

\begin{table}
\centering
\begin{tabular}{lccc}
scenario			&		&\scenario{low}		& \scenario{high} \\
\hline
time to extend		& $T$	&$1\,\millisecond$	&$1\,\second$  \\
speed of extending tip	& $\vC$		&$2.5\,\millimeter/\second$	&$2.5\,\micron/\second$  \\
Reynolds number 	& $\Rey$ 	& $2.5\times10^{-3}$ 	& $2.5\times10^{-10}$\\
Womersley number	& $\Wom$	& $0.03$			& $10^{-5}$ \\
fluid power factor	& $\viscosity \dC \vC^2$	& $6.3\times10^{-3}\,\picowatt$	& $6.3\times10^{-5}\,\picowatt$\\
internal power factor	& $\kfriction \dC^2 \vC^2$	& $6.3\times10^{-3}\,\picowatt$	& $6.3\times10^{-9}\,\picowatt$\\
energy used		& $E$	& $1.5\times10^{-16}\,\joule$	& $1.4\times10^{-15}\,\joule$\\
glucose molecules	& 		& $70$			& $700$\\
\end{tabular}
\caption{Expanding robots for the scenarios of \tbl{scenarios}, with $\dC=L=1\,\micron$, $r=0.75\,\micron$, $n=5$ shells, and $\kfriction=10^3\,\kg/(\meter^2 \second)$ for a robot expanding at a uniform speed from $f=0$ to $f=F=0.5$. The energy listed is for expanding at a uniform rate, given both in joules and the number of glucose molecule oxidations required to produce that energy, at 50\% efficiency~\cite{hogg10}. In both scenarios, fluid viscosity accounts for over $90\%$ of the energy used.}\tbllabel{expanding}
\end{table}

The geometric coefficients, shown in \fig{extending h}, have relatively little variation as the robot expands. This means the uniform expansion rate used in this example is close to the expansion that minimizes the energy. Specifically, the optimal expansion only reduces energy use by about $0.1\%$.

\section{Telescoping Probe}

A major capability of small robots is their ability to access small volumes, e.g., in biological tissue. Nevertheless, some locations may be too small for robots to reach, or the robot may damage surrounding tissue if it enters the region. An alternative is for the robot to extend a narrow probe into the region, e.g., to measure chemicals or release drugs. For instance, sensors, tens of nanometers in size, can be inserted through cell membranes~\cite{duan11,gao12}. These probes are much smaller than the micron-scale robots considered here.

Another application for small probes arises in biological tissues with size-dependent viscosity~\cite{lai09}. In such tissues, micron-size objects experience larger viscosities than smaller ones. In such cases, nano-scale probes could extend through regions of low viscosity that are too small to accommodate the whole robot.

\subsection{Geometry}

\begin{figure}
\centering 
\begin{tabular}{c}
 \includegraphics[width=\figwidth]{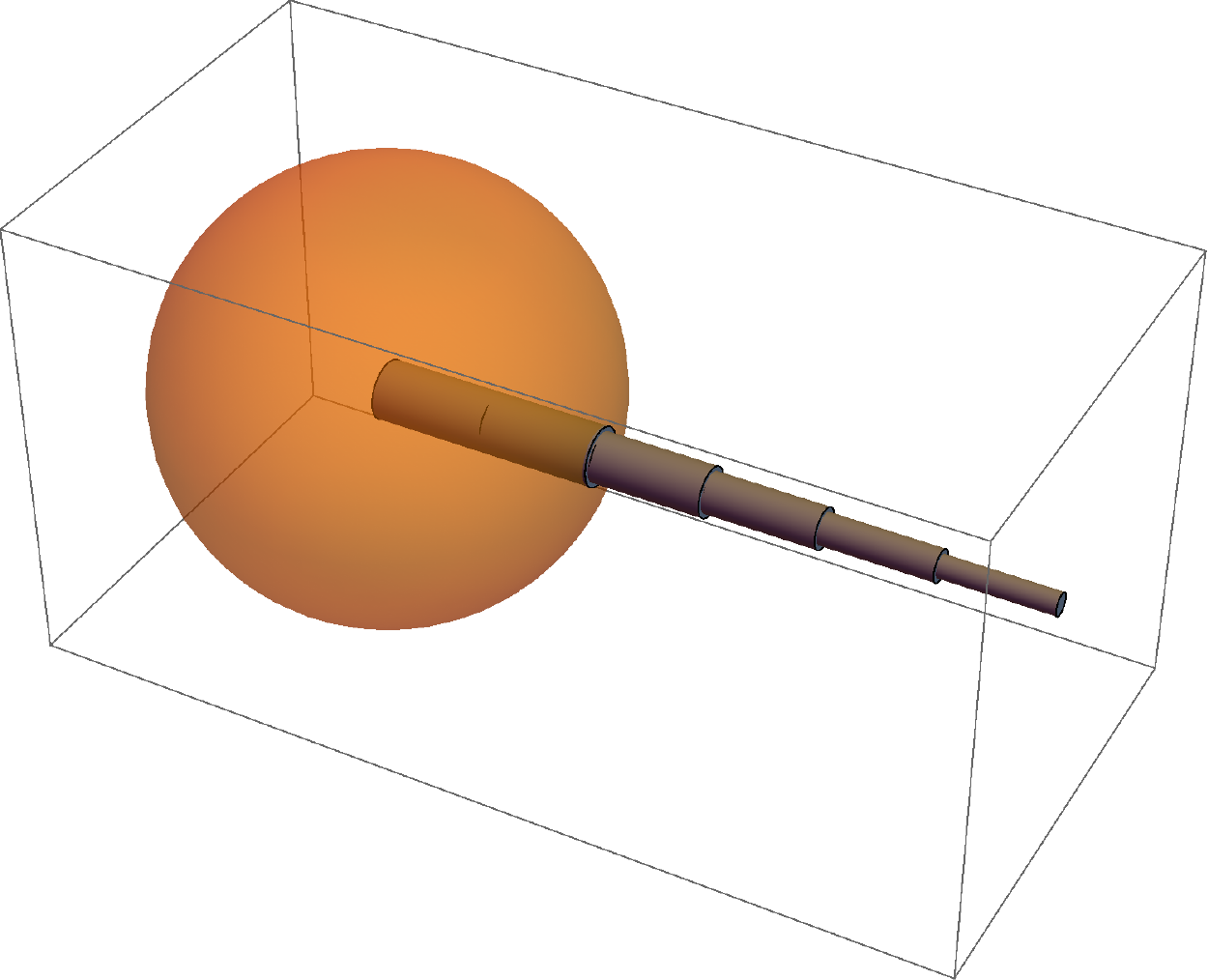}\\
 (a)\\
 \\
  \includegraphics[width=\figwidth]{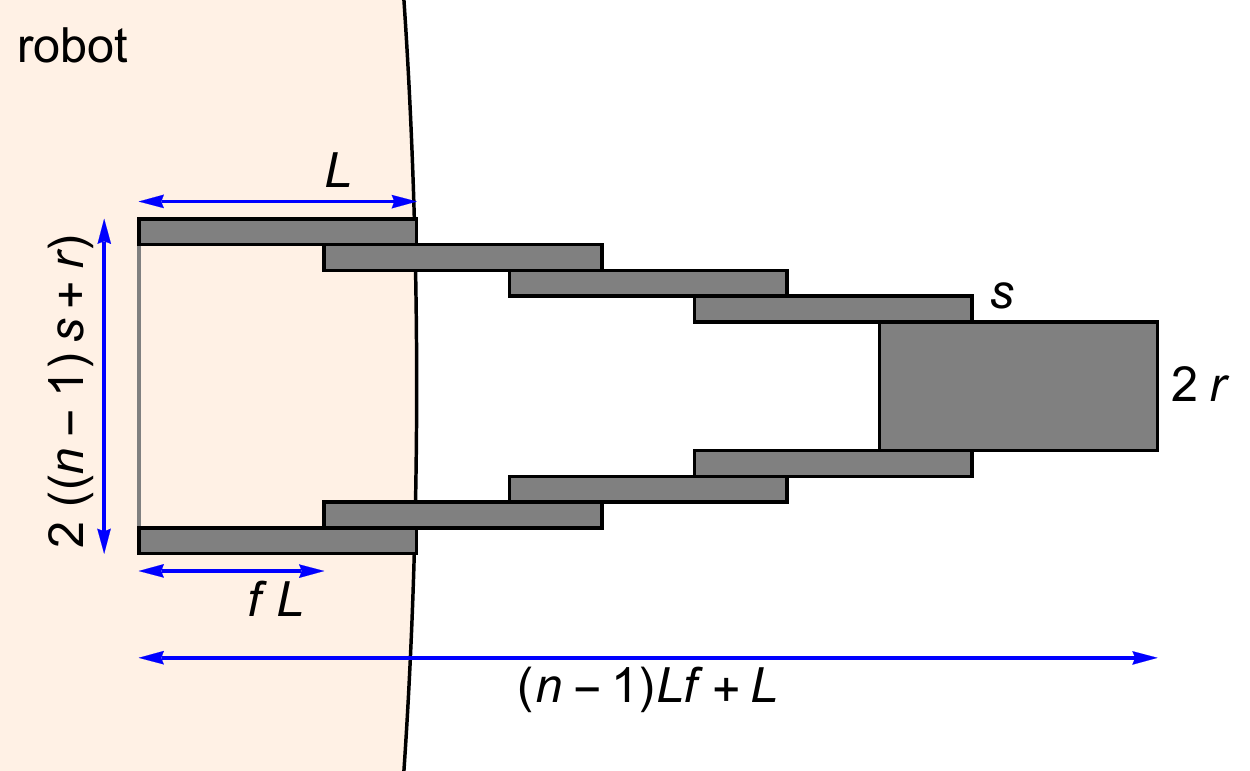}\\
  (b)
\end{tabular}
\caption{Telescoping probe with $n=5$ segments. (a) Probe extending from a spherical robot. (b) Cross section of the probe with an exaggerated vertical scale.
}\figlabel{telescoping probe}
\end{figure}

To focus on the motion of the probe itself, we assume the robot is anchored to a larger structure, so the robot does not move as the probe extends. For simplicity, we consider a probe extending perpendicular to the robot surface, giving the axisymmetric geometry shown in \fig{telescoping probe}.

With $f$ giving the fraction of each segment's extension, we consider uniform extension so $\dot{f}=F/T$, and the probe extends from $f=0$ to $f=F$ in time $T$.
Convenient choices for characteristic size and speed are the segment length and speed of the probe's end:
\begin{eqnarray}
\dC &= & L \\
\vC &=& (n-1) L \dot{f}
\end{eqnarray}

\subsection{Behavior}

We determine $\geometryFluid$ for the telescoping probe using the same simplifications as described above for expanding robots, i.e., consider the external surface of the probe to be truncated cone. The end of the cone moves at speed $(n-1) L \dot{f}$, and the speed along the side interpolates linearly from 0 at the base to $(n-1) L \dot{f}$ at its end.
The result, shown in \fig{telescoping probe h}, shows larger variation than for the expanding robot. This is due to the narrower width of the probe, so a larger proportion of drag arises from the extending sides compared to the constant area of the tip.

\begin{figure}
\centering 
 \includegraphics[width=\figwidth]{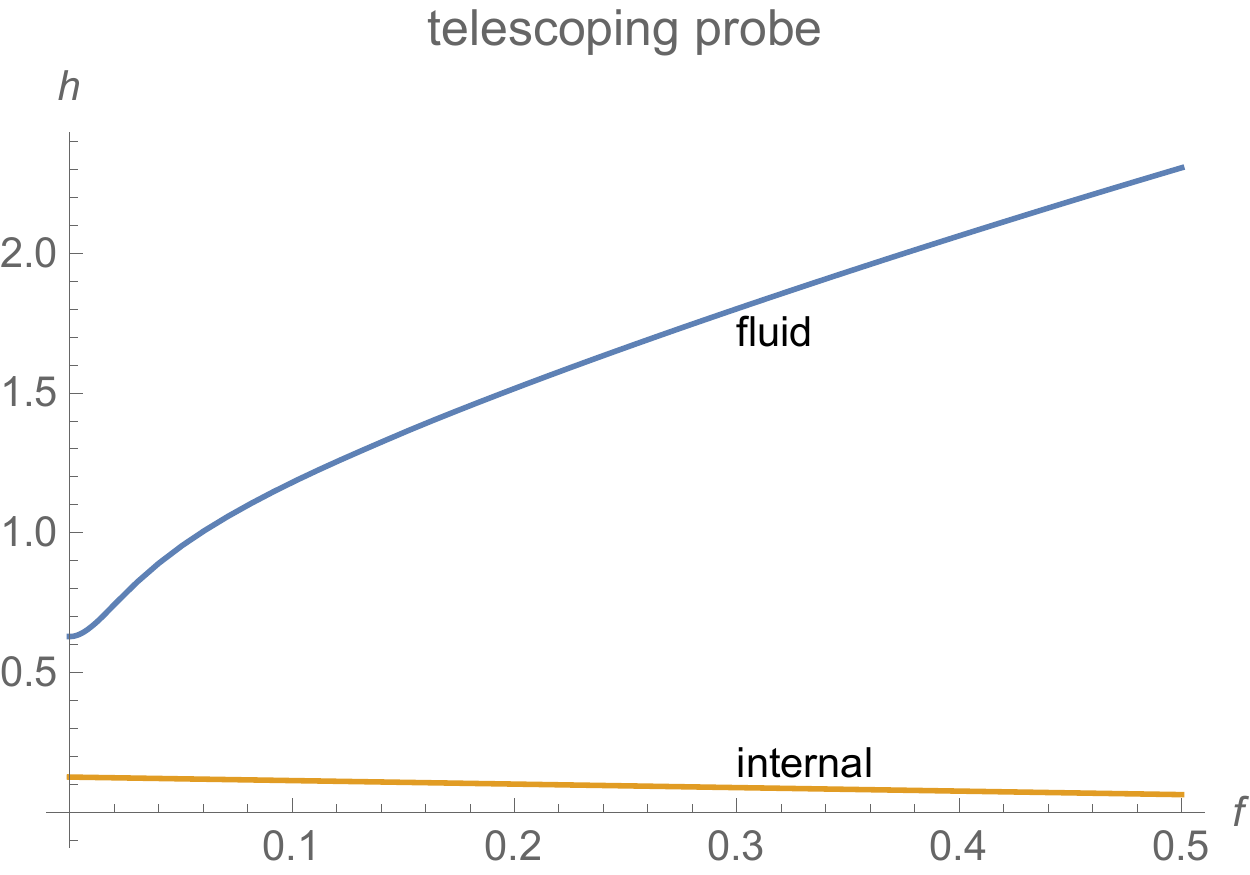}
\caption{Coefficients for viscous drag, $\geometryFluid$, and internal dissipation, $\geometryInternal$, vs. extent of telescoping probe with $n=5$ segments, $r/L=0.05$ and $s/L=0.02$.}\figlabel{telescoping probe h}
\end{figure}

Sliding surface area between segments $i$ and $i-1$ is $S_i= 2\pi(r+(n-i)s) L (1-f)$. The surfaces slide past each other with relative speed $v_i-v_{i-1}=L \dot{f}$.
Internal power use, from the sliding friction between adjacent shells, is similar to that of the expanding robot:
\begin{equation}
\Pinternal = \sum_{i=2}^n \kfriction S_i (L \dot{f})^2
\end{equation}
With \eq{Pinternal}, this gives
\begin{equation}
\geometryInternal = \pi (1-f) \frac{1}{n-1} \left((n-2)\frac{s}{L}+2\frac{r}{L} \right)
\end{equation}
Thus $\geometryInternal$ decreases as the probe expands, i.e., as $f$ increases, since each segment is in contact with less of its neighboring segments. \fig{telescoping probe h} shows the behavior of $\geometryInternal$.

\subsection{Example}

\tbl{telescoping probe} illustrates the situation for the two scenarios of \tbl{scenarios}. The Reynolds and Womersley numbers are small in both cases. 
Combined with the geometric coefficients shown in \fig{telescoping probe h}, these values indicate viscous drag is the dominant contribution to overall power use, and the power required is well below $1\,\picowatt$.

\begin{table}
\centering
\begin{tabular}{lccc}
scenario				&	&\scenario{low}		& \scenario{high} \\
\hline
time to extend		& $T$		&$1\,\millisecond$	&$1\,\second$  \\
speed of extending tip		& $\vC$	&$2\,\millimeter/\second$	&$2\,\micron/\second$  \\
Reynolds number & $\Rey$ 	& $2\times10^{-3}$ 	& $2\times10^{-10}$\\
Womersley number	& $\Wom$	& $0.03$	& $10^{-5}$ \\
fluid power factor	& $\viscosity \dC \vC^2$	& $4\times10^{-3}\,\picowatt$	& $4\times10^{-5}\,\picowatt$\\
internal power factor	& $\kfriction \dC^2 \vC^2$	& $4\times10^{-3}\,\picowatt$	& $4\times10^{-9}\,\picowatt$\\
energy used	& $E$	& $6.8\times10^{-18}\,\joule$	& $6.4\times10^{-17}\,\joule$\\
glucose molecules	& 	& $3$	& $30$\\
\end{tabular}
\caption{Telescoping probe for the scenarios of \tbl{scenarios}, with $\dC=L=1\,\micron$, $r=50\,\nanometer$, $s=20\,\nanometer$, $n=5$ segments, and $\kfriction=10^3\,\kg/(\meter^2 \second)$ for a probe extending at a uniform speed from $f=0$ to $f=F=0.5$. The energy listed is for expanding at a uniform rate, given both in joules and the number of glucose molecule oxidations required to produce that energy, at 50\% efficiency~\cite{hogg10}. In both scenarios, fluid viscosity accounts for over $90\%$ of the energy used.}\tbllabel{telescoping probe}
\end{table}

Minimizing energy requires extending the probe about twice as fast at the beginning of the extension than at the end. This only reduces energy by about 2\% compared to uniform expansion. Of potentially greater significance is balancing power use within the robot: the energy minimizing extension uses power uniformly, at a rate $40\%$ lower than the maximum required during uniform expansion (which occurs as the probe reaches its maximum extent).
Comparing with the smaller improvement for the expanding robot shows that a nonuniform rate of shape change is mainly beneficial for extending long, narrow structures, where the total area exposed to fluid drag changes significantly during the shape change.

\section{Aggregating Robots}

In the above examples, a single robot changes its shape. Another type of metamorphic robot is a modular robot, which is an aggregate of smaller robots (i.e., modules) that move with respect to each other. The aggregate changes shape primarily due to motions of its constituent robots rather than shape changes by those constituents, although they may adjust their shapes to fit into the aggregate. 
This section focuses on the behavior of the smaller robots, i.e., the modules, and takes the aggregate to be a passive structure.

In some cases, modules remain connected in an aggregate as they move~\cite{kotay98,yim00}.
In other cases, more relevant for robots in fluids, the modules can move apart and congregate as needed. Such swarms of separate modules can form aggregates with variable numbers of modules, even with simple controllers and noisy sensors~\cite{arbuckle04,rubenstein14}. This involves navigation for the modules to find each other, e.g., using chemical signals as used by microorganisms~\cite{berg04,dusenbery09} or acoustic signals~\cite{hogg12,santagati14}.
Hydrodynamic interactions could also aid coordination of nearby robots~\cite{riedel05}.
This type of modular robot introduces a new ingredient not found in the previous examples: robots moving through the fluid toward each other. 

For simplicity, we consider robot motion in an otherwise stationary fluid and take the robots to be neutrally buoyant spheres. This discussion does not include dissipation associated with any locking mechanisms when modules attach to the growing aggregate, or arising from small shape adjustments, if any, the modules require for this attachment.

\subsection{Geometry}

To form an aggregate, the robots propel themselves toward each other through the fluid. Power dissipation depends on their locomotion method. Methods involving extended structures, such as flagella, pose a problem of tangling as the robots approach each other. To avoid this problem, we suppose robots use tangential surface motion for propulsion~\cite{leshansky07}. 
Specifically, we consider treadmills in an equatorial band covering half the robot surface area~\cite{hogg14}. With this method, the robot does not change shape as it moves, so internal power dissipation depends on the speed of the treads but not on the position of the robot relative to other robots.

In the previous examples, robot actuators directly altered the shape of the robot, characterized by the parameter $f$. In the case of modular robots, the configuration is the relative positions of the modules. Actuators do not directly specify the configuration. Instead, actuators exert force against the fluid to move the robot. The speed of this motion depends on the positions of other robots. Thus there is not a fixed relationship between actuator motion and configuration of the aggregate. This means robots moving in fluid are a more complicated shape-changing situation, with 
two properties determining the behavior, i.e., the relations of tread speed to locomotion speed, and of tread speed to fluid dissipation.

Convenient choices for characteristic size and speed are the sphere radius and tread speed over the sphere surface:
\begin{eqnarray}
\dC &= & a \\
\vC &=& \vTread
\end{eqnarray}

In addition to characterizing power dissipation (\eq{Pfluid} and \eq{Pinternal}), the tread speed  determines the sphere's locomotion speed $u$ according to~\cite{hogg14}
\begin{equation}\eqlabel{locomotion}
u = \geometryLocomotion \vTread
\end{equation}
where the proportionality $\geometryLocomotion$ varies with the sphere's position.

To illustrate the behavior of modular robots, we consider two cases, shown in \fig{sphere}.
First, a sphere approaches a much larger, stationary aggregate, modeled as a sphere approaching a planar wall. 
In the second case, two identical spheres approach each other along their common axis. This illustrates the behavior at an early stage of aggregation, i.e., when the first two robots approach each other to start forming an aggregate.

We characterize the position of the sphere by the distance $L$ of the center of the sphere to the wall, in the first case, and from the center of one sphere to the midpoint between the two spheres, in the second case, so the distance between their centers is $2L$. We define the normalized distance
\begin{equation}\eqlabel{delta}
\delta = L/a
\end{equation}
so $\delta=1$ corresponds to the sphere touching the wall or the other sphere in the two cases, respectively.

\begin{figure}
\centering 
\includegraphics[width=\figwidth]{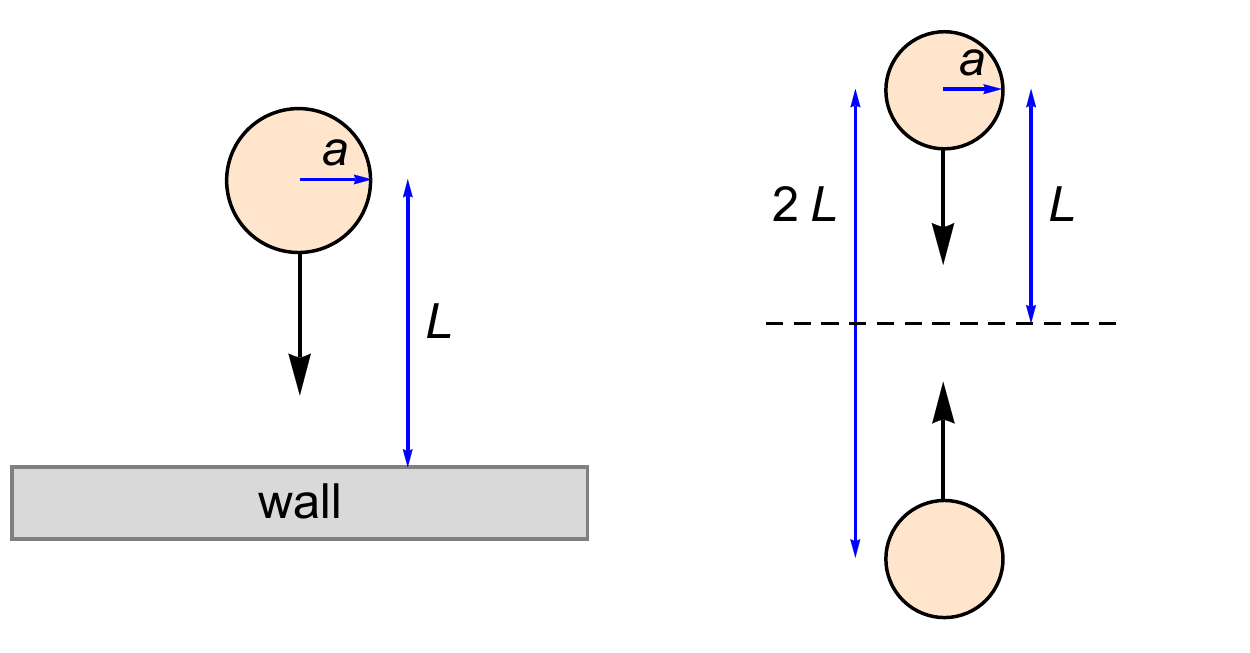}
\caption{Cross section of a sphere approaching a wall or another sphere.}\figlabel{sphere}
\end{figure}

\subsection{Behavior}

\begin{figure}
\centering 
 \includegraphics[width=\figwidth]{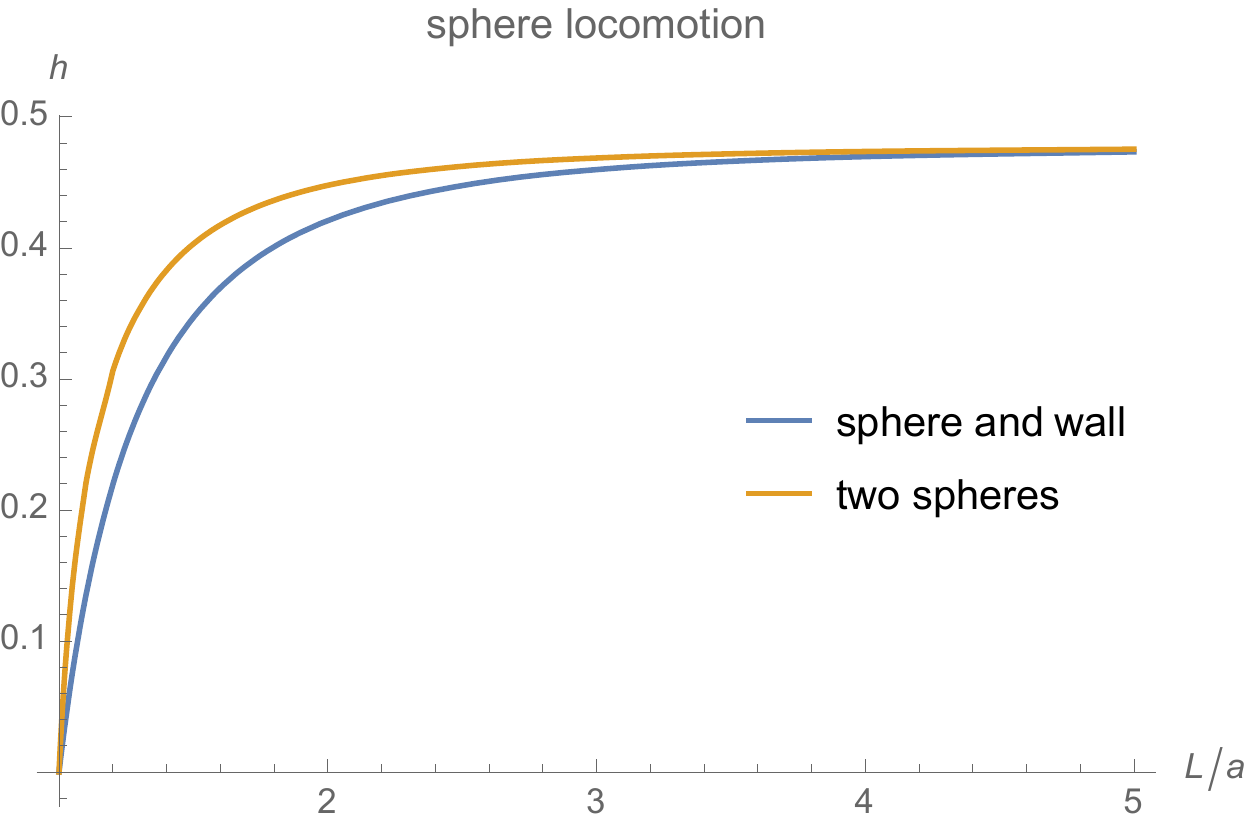}
\caption{Ratio of locomotion speed to tread speed, \geometryLocomotion, vs.~normalized distance $\delta=L/a$ (\eq{delta}) for a sphere approaching a wall and for two spheres approaching each other.
}\figlabel{sphere h-locomotion}
\end{figure}

\begin{figure}
\centering 
\includegraphics[width=\figwidth]{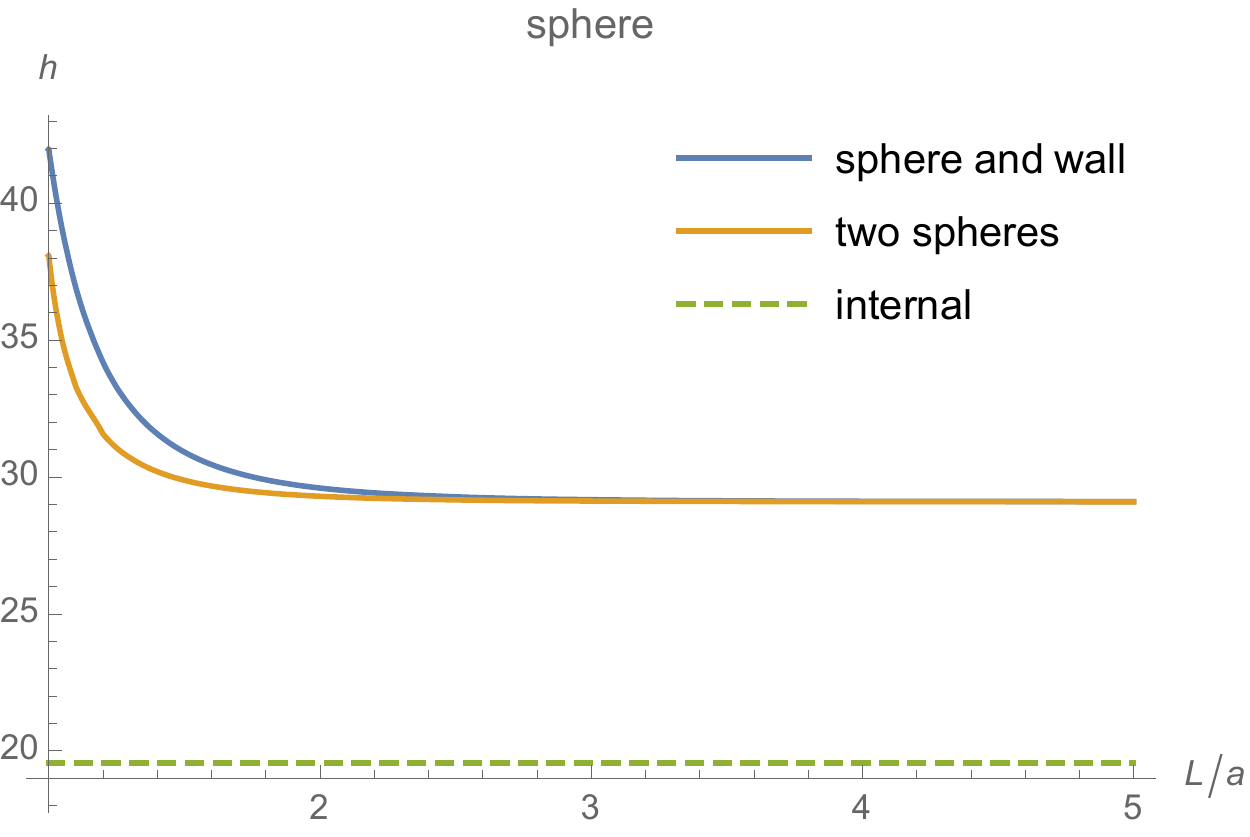}
\caption{Coefficients for viscous drag, $\geometryFluid$ (solid), and internal dissipation, $\geometryInternal$ (dashed), vs.~normalized distance $\delta=L/a$ (\eq{delta}) for a sphere approaching a wall and for two spheres approaching each other.
Internal dissipation is the same for both cases and independent of distance.}\figlabel{sphere h}
\end{figure}

We determine $\geometryLocomotion$ and $\geometryFluid$ for the sphere by assuming quasi-static Stokes flow. 

\fig{sphere h-locomotion} shows $\geometryLocomotion$ for the two cases. As the separation decreases toward zero (i.e., $\delta \rightarrow 1$), the treadmill locomotion becomes increasingly less effective, i.e., a given tread speed produces slower motion through the fluid.
With two approaching spheres, the distance between them decreases at twice the speed of each sphere with respect to the fluid.

We consider a situation where the robots start relatively far apart, with $\delta=\delta_0$ somewhat larger than one, and move toward each other to a smaller distance $\delta_1\approx 1$. Since $\geometryLocomotion \rightarrow 0$ as $L \rightarrow a$,
we consider motion of robots to within a small positive distance, e.g., $50\,\nanometer$, which is sufficient for the robots to link to each other.
This corresponds to $\delta_1=1.05$ for a $1\,\micron$-radius sphere approaching a wall, and $\delta_1=1.025$ for two spheres approaching each other.

\fig{sphere h} shows the power dissipation: values for $\geometryFluid$ and $\geometryInternal$ for the two cases. The internal dissipation coefficient is independent of distance.
When the distance is a few multiples of the sphere radius, fluid power dissipation is close to that of an isolated sphere using this tangential motion.\footnote{The dissipation shown here is somewhat larger than the value ignoring fluid vorticity~\cite{hogg14}, due to additional dissipation arising from the velocity gradient at the ends of the treadmill. Dissipation due to this edge-effect depends on the distance at the ends of the tread over which velocity changes. This study uses a distance corresponding to $50\,\nanometer$ bearings for the treadmill~\cite{hogg14}.}

\subsection{Example}

\begin{table}
\centering
\begin{tabular}{lccc}
scenario				&	&\scenario{low}		& \scenario{high} \\
\hline
tread speed		& $\vC$	&$1\,\millimeter/\second$	&$10\,\micron/\second$  \\
Reynolds number & $\Rey$ 	& $10^{-3}$ 	& $10^{-9}$\\
Womersley number	& $\Wom$	& $0.03$	& $3\times10^{-5}$ \\
fluid power factor	& $\viscosity \dC \vC^2$	& $10^{-3}\,\picowatt$	& $10^{-3}\,\picowatt$\\
internal power factor	& $\kfriction \dC^2 \vC^2$	& $10^{-3}\,\picowatt$	& $10^{-7}\,\picowatt$\\
%
diffusion coefficient	& $\BoltzmannConstant \Tbody/(6\pi\viscosity \aSphere)$	& $10^{-13}\,\meter^2/\second$	& $10^{-17}\,\meter^2/\second$\\
\end{tabular}
\caption{Sphere motion for the scenarios of \tbl{scenarios}, with $\dC=\aSphere=1\,\micron$ and $\kfriction=10^3\,\kg/(\meter^2 \second)$ for a sphere with uniform tread speed in an equatorial band covering half the surface area. $\BoltzmannConstant$ is the Boltzmann constant and $\Tbody=310\,\Kelvin$ is body temperature.
}\tbllabel{sphere}
\end{table}

\tbl{sphere} illustrates the behavior, including diffusion coefficients of spheres~\cite{berg93}, for the scenarios of \tbl{scenarios}. 
\tbl{sphere cases} gives the time and energy used for the two scenarios. 
Over these times, the diffusion coefficients given in \tbl{sphere} indicate the effect of Brownian motion: moving the spheres by typical distances $\sqrt{6 D T}$ of about $100\,\nanometer$ and $10\,\nanometer$, for the \scenario{low} and \scenario{high} scenarios, respectively. These distances are small compared to the range of motion considered here, i.e., several microns. Thus, Brownian motion is noticeable but fairly minor for the scenarios considered here.

The tread speeds in \tbl{sphere} and geometric coefficients in \fig{sphere h} indicate viscous drag is the dominant contribution to overall power dissipation, which is well below $1\,\picowatt$ for the times in \tbl{sphere cases}.
The dissipation is per sphere, so total power in two-sphere case is twice the value given here. 
The relatively small variation in dissipation as a function of position (\fig{sphere h}) leads to only a slight reduction in energy use ($0.1\%$ or less) when optimizing the tread speed as the robot moves.

\begin{table}
\centering
\begin{tabular}{lccc|cc}
scenario				&	&\multicolumn{2}{c}{\scenario{low}}		& \multicolumn{2}{c}{\scenario{high}} \\
case				&	&wall	&two spheres	& wall	& two spheres \\
\hline
final normalized distance	& $\delta_1$	& $1.05$	& $1.025$	& $1.05$	& $1.025$ \\
motion time	& T	& $10\,\millisecond$ 	&$9\,\millisecond$	& $1\,\second$ 	&$0.9\,\second$ \\
energy used	& $E$	& $4.9\times10^{-16}\,\joule$	& $4.6\times10^{-16}\,\joule$	&$3.0\times10^{-14}\,\joule$	&$2.8\times10^{-14}\,\joule$\\
glucose molecules	& 	& $250$	& $230$	& $15000$	& $14000$\\
\end{tabular}
\caption{Sphere motion for the scenarios of \tbl{sphere}, starting from normalized distance $\delta_0=5$ with uniform tread speed.
Each scenario shows behavior for two cases: a sphere approaching a wall and two spheres approaching each other. 
The energy listed is given both in joules and the number of glucose molecule oxidations required to produce that energy, at 50\% efficiency~\cite{hogg10}. 
In the \scenario{low} and \scenario{high} scenarios, fluid viscosity accounts for $60\%$ and nearly $100\%$ of the energy used, respectively.}\tbllabel{sphere cases}
\end{table}

An additional issue for aggregating robots arises if they obtain power from chemicals in their environment. In that case, as robots approach there is less power available due to competition for those chemicals from other robots~\cite{mcdonald05}. 
This could increase the time required to aggregate many robots in small volumes.
However, this is not a significant issue for modest numbers of robots obtaining power from oxygen and glucose dissolved in the fluid. This is because, for the scenarios in \tbl{sphere cases}, the energy requirements are considerably less than available chemical power for a few dozen robots operating in close proximity in or near blood vessels~\cite{hogg10}.

\section{Discussion}

Microscopic metamorphic robots with the parameters studied here will likely have enough power from chemicals available in biological tissues~\cite{freitas99,hogg10}. Biological cells indicate this is physically possible, since cells actively change their shapes in spite of facing similar power limitations and dominance of surface forces as the robots considered in this paper.

There are significant engineering challenges to creating these robots. These include building robots from materials whose sliding friction is as low as assumed with \eq{k friction}. With this value, dissipation from internal friction and external viscous drag are of comparable magnitude for the examples considered in this paper. Thus there is is relatively little decrease in power use from additional reduction in friction.
On the other hand, building the robots from conventional materials, which have much larger friction, will lead to significantly more power dissipation, almost all of which will be due to internal friction.

The focus in this paper is on shape change involving a single degree of freedom. More complex changes include extending multiple probes or manipulator arms. 
Similarly, modular robots can involve several modules moving simultaneously on or near the surface of the aggregate~\cite{bojinov02,kotay98a,rubenstein14}.
Simultaneous moving parts lead to complex hydrodynamic interactions, similar to those for nearby objects moving in a fluid~\cite{hernandez05,kim05,lauga09}.

Extensions to this study could examine other types of reconfiguration. For instance, because surface forces dominate behavior of microscopic robots, changing surface properties such as affinity for water~\cite{lahann03} could complement shape changes in allowing robots to adapt to their tasks.

A final direction for study is developing controls suited to microscopic metamorphic robots, particularly their limited computation and noisy sensors. One approach, for modular robots, is extending simple local rules that grow structures~\cite{bojinov02,salemi01,werfel14} and other swarm algorithms~\cite{bonabeau99} to account for large surface forces and Brownian motion.
Automated rule generation~\cite{kubica02} could complement these approaches.


\end{document}